\documentclass{article}

\usepackage{microtype}
\usepackage{graphicx}
\usepackage{subfigure}
\usepackage{booktabs} 

\usepackage{hyperref}

\usepackage[accepted]{mlsys2026}

\usepackage{amsmath,amssymb}
\usepackage{mathtools}
\usepackage{amsthm}
\usepackage{derivative}
\usepackage[normalem]{ulem}

\usepackage{booktabs}
\usepackage{multirow}
\usepackage{caption}
\usepackage{subcaption}
\usepackage{xcolor}
\usepackage{colortbl}
\definecolor{mylightyellow}{rgb}{1, 1, 0.6}
\definecolor{lightyellow}{rgb}{1.0, 1.0, 0.88}
\definecolor{azure(web)(azuremist)}{rgb}{0.94, 1.0, 1.0}

\usepackage{microtype}
\usepackage{graphicx}

\usepackage{eqparbox,array}
\usepackage{algorithm}
\renewcommand{\algorithmicrequire}{\textbf{Input:}}
\renewcommand{\algorithmicensure}{\textbf{Output:}}
\renewcommand{\algorithmiccomment}[1]{\hfill$\triangleright$ \textit{#1}}

\usepackage[capitalize,noabbrev]{cleveref}

\usepackage{url}

\usepackage{listings}
\expandafter\def\expandafter\normalsize\expandafter{%
    \normalsize
    \setlength\abovedisplayskip{2pt}
    \setlength\belowdisplayskip{2pt}
    \setlength\abovedisplayshortskip{2pt}
    \setlength\belowdisplayshortskip{2pt}
}

\usepackage{enumitem}

\theoremstyle{plain}

\theoremstyle{definition}

\theoremstyle{remark}

\usepackage[textsize=tiny]{todonotes}

\definecolor{darkgray}{rgb}{0.66, 0.66, 0.66}

\newcommand{\cb}{}  

\newcommand{\cgray}{\textcolor{gray}}

\newcommand{\round}[1]{\ensuremath{\lfloor#1\rceil}}

\def \SYS{MixLLM}

\begin{document}

\twocolumn[
\mlsystitle{\SYS{}: LLM Quantization with Global Mixed-precision between Output-features and Highly-efficient System Design}




\begin{mlsysauthorlist}
\mlsysauthor{Zhen Zheng}{msft}
\mlsysauthor{Xiaonan Song}{msft}
\mlsysauthor{Chuanjie Liu}{msft}
\end{mlsysauthorlist}

\mlsysaffiliation{msft}{Microsoft}

\mlsyscorrespondingauthor{Zhen Zheng}{zhengzhen@microsoft.com}


\vskip 0.3in

\begin{abstract}

Quantization has become one of the most effective methodologies to compress LLMs into smaller size.
However, the existing quantization solutions still show limitations of either non-negligible accuracy drop or low system efficiency.
In this paper, we propose \SYS{} that explores the optimization space of mixed-precision quantization between output features, based on the insight that different features matter differently in the model.
\SYS{} identifies the important output features in the global view rather than within each single layer,
effectively assigning larger bit-width to output features that need it the most to achieve high accuracy and low memory usage.
We present the sweet spot of quantization configuration of algorithm-system co-design with high accuracy and system efficiency.
To address the system challenge, we design the two-step dequantization to make use of the Tensor Core easily and fast data type conversion to reduce dequantization overhead, and present the software pipeline to overlap the memory access, dequantization and the MatMul to the best.
Extensive experiments show that with only 10\% more bits, the perplexity increase can be reduced from about 0.5 in SOTA to within 0.2 for Llama 3.1 70B, while MMLU-Pro loss can be reduced from 1.92 to 0.99 over the SOTA of three popular models.
Besides its superior accuracy, \SYS{} also achieves state-of-the-art system efficiency.
\cb{Code is released at \url{https://github.com/microsoft/MixLLM}.}


\end{abstract}
]



\printAffiliationsAndNotice{}  

\section{Introduction}
\label{sec:introduction}

Large language models (LLMs)~\cite{exp-on-gptq4,llama3} have shown remarkable performance on various tasks.
But their large memory consumption and massive computation cost have become an obstacle for the efficient deployment~\cite{flash-llm,quant-llm}.
Quantization has become one of the most effective solution to compress LLMs into smaller size~\cite{gptq,awq,smoothquant,zeroquant},
by representing the weight or activation with smaller bit-width.
However, the existing quantization solutions still show limitations of either non-negligible accuracy drop or system inefficiency.

There is a triangle of characteristics for LLM quantization: \textit{accuracy}, \textit{memory consumption} of parameters, and \textit{system efficiency} of execution, 
The existing quantization solutions have different focus and trade-off in the triangle:
1) The weight-only method targets to solve the memory consumption problem, and can speedup the small-batched decoding execution that faces the memory-wall problem~\cite{flash-llm,squeezellm}.
But their accuracy drop of 4-bit quantization can be a challenge for the production workloads sensitive to accuracy, 
as illustrated in recent studies~\cite{zeroquant-fp6, kumar2024scaling}.
Besides, the weight-only method can lead to system performance drop for large-batched workloads due to the dequantization overhead.
2) The weight-activation quantization represents the activation with low-bit values along with the weights, potentially lead to higher system efficiency.
But it can lead to larger accuracy drop than the weight-only method as the activation is usually harder to quantize~\cite{atom,quarot,qserve}.
Besides, it introduces more dequantization overhead for the activation that can hurt the system efficiency.

\begin{figure*}
    \centering
    \includegraphics[width=.8\textwidth]{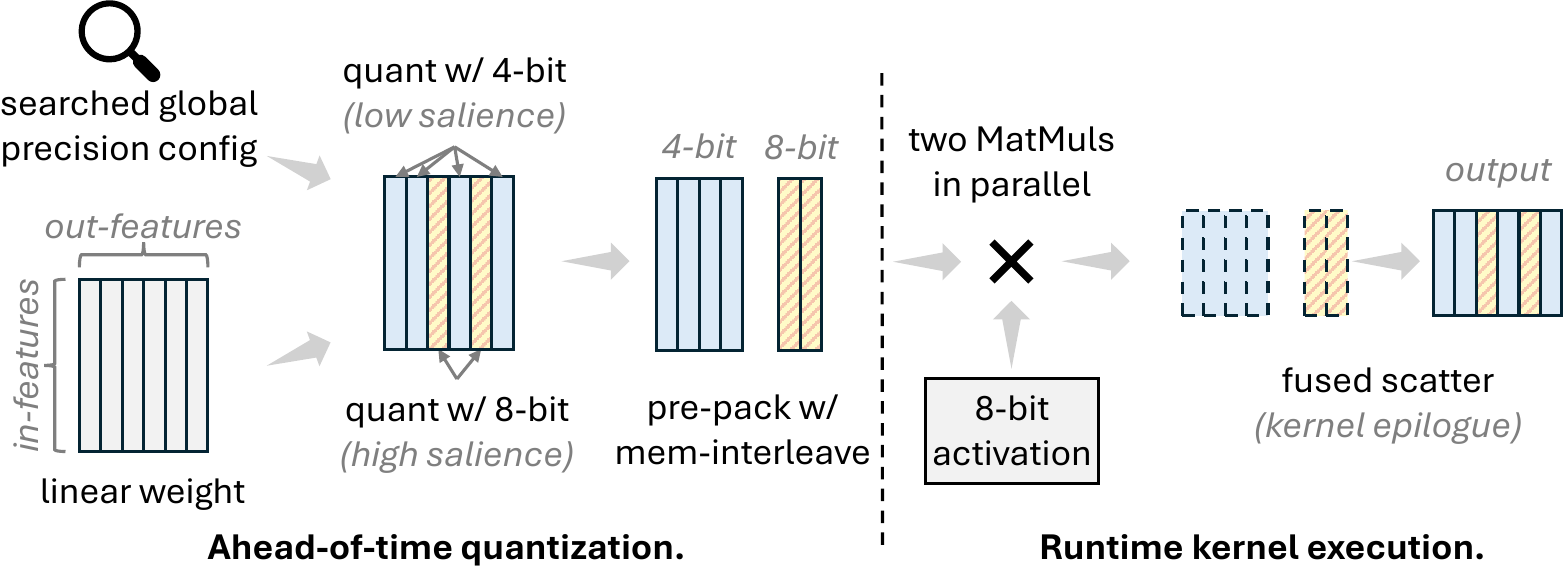}
    \caption{Illustration of the quantization with mixed-precision between output features and kernel execution.}
    \label{fig:mixed-precision-linear}
\end{figure*}

\textbf{Contributions.}
In this paper, we provide an extensive
analysis of the general quantization principles, and propose \SYS{} with the following contributions:

\textbf{$\blacktriangleright$ High accuracy with low memory consumption: mixed-precision between output features on the weight, with global salience identification.}
Given that different neurons matter differently to the model's output, we use different bit-width for different output features (i.e., output channels) for the weight quantization (Fig.\ref{fig:mixed-precision-linear}).
Rather than using a uniform number of outliers within each layer according to the estimated salience w.r.t. each single layer~\cite{atom, slim-llm}, \SYS{} identifies the salience of different output features globally according to the estimated loss to the model's output.
This is because different layers can have different importance to the model.
Besides, the mixed-precision between output features makes the system design easier than between input features because the calculation of different output features are disjoint sub-problems.

\textbf{$\blacktriangleright$ High accuracy with good system efficiency: the co-designed quantization configuration and GPU kernel optimization.}
We observe the sweet spot of several quantization decisions to achieve both good accuracy and system efficiency.
\SYS{} uses 8-bit for activation quantization as it can retain a good accuracy.
Besides, MatMul execution tends to be bound more on the larger weight tensor rather than the smaller activation tensor, which weakens the need to push the activation smaller (refer to Sec.\ref{subsec:decision}).
\SYS{} uses symmetric quantization for 8-bit and asymmetric for 4-bit for good accuracy, both in group-wise manner.
Such configuration makes it challenging to achieve good system efficiency.
We design the two-step dequantization to enable using fast int8 Tensor Core for such configuration,
along with the fast integer-float conversion to reduce the dequantization overhead.
We make the software pipeline design to tackle the system challenge of group-wise and 4-bit asymmetric quantization, 
and present the state-of-the-art group-wise int8 quantization GPU kernel, supporting asymmetric 4-bit and symmetric 8-bit weight.
\cb{We also provide the quantization-aware Roofline analysis in this paper.}

Extensive evaluation shows that, with 10\% 8-bit and 90\% 4-bit (i.e., W4.4A8), \SYS{} significantly outperforms SOTA quantization algorithms while achieving the state-of-the-art system efficiency \cb{on both A100 and H100 GPUs}.
\section{Background, Related Work, and Discussion}
\label{sec:related}


Quantization maps the tensor $X$ into the target range with smaller bit-width through $X_q = clamp(\round{\frac{X}{s}} + z, range)$, where $s$ is the scale and $z$ is the zero point.
The scale and zero point can be calculated from the whole channel/token vector or a smaller group.
The group-wise scheme results in higher accuracy but requires more complex GPU kernel design.
The symmetric quantization uses 0 as the zero point, which simplifies the computations 
and enables many works to design the per-channel/per-token quantized kernels by multiplying the scales at the epilogue of the whole MatMul (matrix multiplication) for dequantization~\cite{smoothquant, trtllm}.
However, it usually leads to larger loss than the asymmetric one, especially for smaller bit-widths like 4-bit.

\subsection{Related Work and Discussion}
\label{subsec:analysis}

This paper focuses on pure post-training quantization (PTQ), without the training on the weight (i.e., QAT) or the quantization parameters (e.g., training the rotation matrix like in SpinQuant~\cite{spinquant}).

\textbf{Systems that affect the quantization requirement.}
The continuous batching technology~\cite{orca} enables to batch the decoding tasks from different requests together to enlarge the batch dimension of MatMul during LLM inference.
The chunked-prefill method~\cite{splitfuse, sarathi-serve, zheng2024batchllm} advances the continuous batching by merging the prefill and decoding tasks into the same batch, further enlarging the MatMul shapes.
These technologies pushes many LLM jobs to become compute-bound and motivate the demand to reduce computation.

\textbf{Weight-only quantization and its limitation.}
There emerges a wide range of technologies to improve the accuracy of weight-only quantization.
GPTQ~\cite{gptq} advances OBC~\cite{obc} on OBS-based~\cite{obs} weight compensation with blocked updating and reordering.
AWQ~\cite{awq} proposes to scale the weight according to the characteristic of activation.
OmniQuant~\cite{OmniQuant} proposes the learnable scaling and weight clipping factors.
SpQR~\cite{spqr}, SqueezeLLM~\cite{squeezellm} and OWQ~\cite{owq} separate the outliers from the quantiation and with half precision.
QuIP~\cite{quip} and QuIP\#~\cite{quip-sharp} aim to achieve extremely low-bit quantization with incoherence processing, QTIP~\cite{qtip} also leverages incoherence processing.
AQLM~\cite{aqlm} and PV-Tuning~\cite{PV-Tuning} target extremely low-bit quantization.
SliM-LLM~\cite{slim-llm} focuses on extremely low-bit quantization using intra-layer mixed-precision.
ZeroQuant(4+2)~\cite{zeroquant-fp6} and Quant-LLM~\cite{quant-llm} aim to improve accuracy with FP6 quantization.

The weight-only quantization does not reduce the computation but introduces the extra dequantization operations, as it requires to dequantize the weight into float16 for execution.
The current weight-only quantization faces two challenges:
1) From the accuracy aspect, there is still an accuracy gap between the 4-bit quantization and the float16 model, especially for many real business scenarios sensitive to the small accuracy drop, as discussed in the recent works~\cite{zeroquant-fp6, quant-llm}.
2) It can lead to system efficiency drop on busy servers as the recent LLM inference serving systems will usually batch the processing of different requests together on the server and form large MatMuls.
The large MatMuls are compute-bound and will suffer from the dequantization overhead~\cite{qserve}.

\textbf{Weight-activation quantization and the challenges.}
The weight-activation quantization helps to make use of the low-bit computing unit.
LLM.int8()~\cite{llm-int8} observes the activation outlier problem and separates outliers from quantization with half precision.
ZeroQuant~\cite{zeroquant} proposes the per-token activation quantization and group-wise weight quantization.
SmoothQuant~\cite{smoothquant} addresses the activation outlier problem through smoothing,
and AffineQuant~\cite{affinequant} proposes the general affine transformation for quantization.
RPTQ~\cite{rptq} reorders the channels to cluster similar scaled values together.
SpinQuant~\cite{spinquant} and QuaRot~\cite{quarot} leverages matrix rotation properties to alleviate the outlier phenomenon.
Atom~\cite{atom} uses the mixed-precision between input features to improve accuracy of 4-bit activation quantization.
QServe~\cite{qserve} is a holistic solution of W4A8 quantization.

Even though the weight-activation quantization has the advantage of reduced MatMul computation (i.e., smaller bit-width computation with higher throughput), it faces the challenge of accuracy drop caused by the activation quantization.
The SOTA low-bit weight-activation solutions~\cite{quarot, spinquant, qserve} still have a gap to the 4-bit weight only quantization.
Besides the accuracy drop, the activation quantization will introduce more dequantization overhead than the weight-only one, which makes it challenging to design efficient GPU kernels.

When enabling the asymmetric quantization, the result of $(X_q - z)$ may exceed the range of the bit-width of $X_q$, making it hard to use the corresponding Tensor Core computation.
Systems like Atom~\cite{atom} thus avoid using the asymmetric quantization, with the cost of larger accuracy drop.
The integer quantization requires integer-to-float (I2F) conversion to apply scales.
However, the I2F instruction is more expensive than the common computations on modern GPUs~\cite{demystifying-ampere} and can lead to large system performance drop for group-wise quantization ($>10\%$ drop in our practice).
Besides, the throughput of Tensor Core is much higher than SIMT Cores, 624 TOPS of int8 Tensor Core vs. 19.5 TFLOPS/TOPS of FP32/INT32 SIMT Cores on A100 GPU.
There lacks a well designed software pipeline to overlap the Tensor Core computation and SIMT Core based dequantization for the group-wise and asymmetric weight-activation quantization.


\textbf{Performance challenge of the float16 outlier separation}
Outlier separation with half precision works to improve the accuracy while using small bit-width for the non-sensitive weights~\cite{squeezellm, spqr}, by separating the outliers into an extra sparse tensor in float16.
However, it is hard to achieve the peak performance due to the inefficiency of the sparse computation on the GPU, especially when the batch size is large and the linear layer becomes compute-bounded.
(As discussed in Flash-LLM~\cite{flash-llm}, the hardware utilization can be lower than 10\% for the sparse MatMul, while its dense counterpart can usually achieve more than 60\%.)
This is because the unstructured tensor computation cannot make use the fast Tensor Core easily, but has to use the SIMT Core in float16 for computation and float32 for accumulation\footnote{Flash-LLM~\cite{flash-llm} optimizes the unstructured sparse MatMul, but can only speedup the small-batched scenarios.}.
Moreover, sparse computing makes it more difficult to fully utilize the hardware due to the non-continuous memory pattern and the extra index computation.

\section{Methodology}
\label{sec:mixed}

\subsection{Quantization Design and Decision in \SYS{}}
\label{subsec:decision}

We make the following design and decision to optimize the quantization algorithm and system.


\subsubsection{Global Mixed-precision}

Different elements of the weight show different salience to the network's loss when being quantized~\cite{squeezellm, spqr}.
The outlier separation method can improve the accuracy by using float16 for high-salience elements, but can suffer from the inefficient sparse MatMul.
We observe that the elements with high salience tend to show distribution along the output channels for most of the linear layers in many LLMs.
Based on this observation, we can assign larger bit-width to the output channels of high salience, and smaller bit-width to the others.
Through the experiments, we get the same conclusion with the existing works~\cite{squeezellm, spqr} that there is only a small set of elements with high salience to quantization.
Thus we only need to assign the large bit-width to a small portion of the output channels to achieve good accuracy while retaining a small memory consumption.

The structured mixed-precision between output channels can be friendly to the system efficiency and kernel development,
due to the nature that different output features are disjoint in the MatMul and the computation of them are different sub-problems.
Fig.\ref{fig:mixed-precision-linear} illustrates the mixed-precision quantization and execution in \SYS{}.
It divides the linear into independent sub-problems, and finally gathers the output of the sub-problems together to form the result.
This optimization space is orthogonal to the existing quantization methodologies and can be applied together with them.

\begin{figure*}
    \centering
    \includegraphics[width=\textwidth]{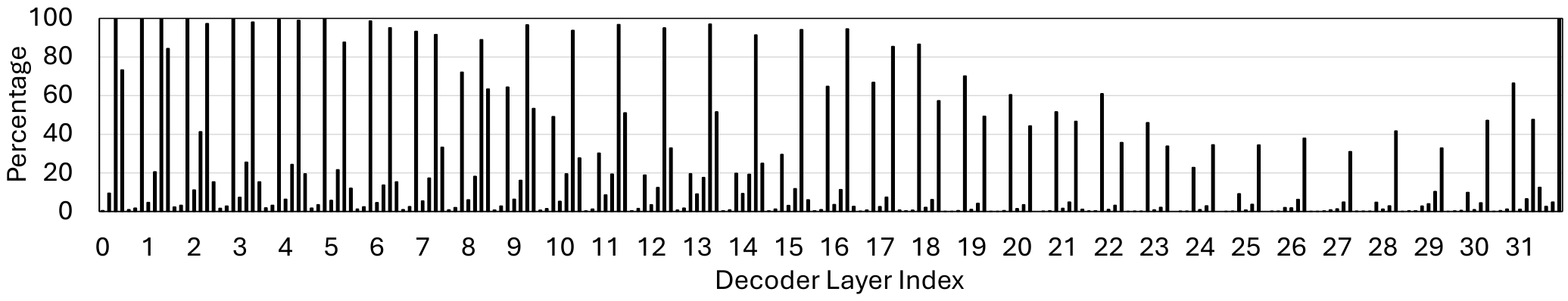}
    \caption{The percentage of high-salient out features within each linear layer of Llama 3.1 8B model according to each feature's contribution to the final loss after quantizing to 4-bit, with 10\% high-salient features globally.
    Each decoder layer contains \texttt{q\_proj}, \texttt{k\_proj}, \texttt{v\_proj}, \texttt{o\_proj}, \texttt{gate\_proj}, \texttt{up\_proj}, and \texttt{down\_proj} in order.}
    \label{fig:percentage-of-8bit}
\end{figure*}

One critical problem is how to identify the high-salience output channels in the model.
The fixed threshold~\cite{spqr} or the fixed number/ratio~\cite{atom, owq} of high salience elements computed by the local loss of layers can be sub-optimal to the end-to-end model,
as different layers show different importance to the model's final output~\cite{ineffectiveness,shortgpt,hawq}.
A high salience channel w.r.t. a layer may not be a high salience channel of the end-to-end model.
In \SYS{}, we compute the high salience channels globally according to their impact to the model's final loss (Sec.\ref{subsec:search}).
As a result, different layers will have different number of high salience channels.
Fig.\ref{fig:percentage-of-8bit} shows the distribution of the top 10\% high-salient out features in Llama 3.1 8B, showing huge difference between different layers.

Note that this design is different from the mixed-precision in Atom~\cite{atom} from two aspects.
1) \SYS{} first addresses the problem of identifying the high-salience channels globally rather than locally.
2) \SYS{} applies the mixed-precision between output features rather than input features, which is more system/algorithm flexible as the output features are disjoint naturally.
It also differs from SliM-LLM~\cite{slim-llm} which also only considers the local loss to determine the mixed precision and does not focus on the system performance problem.

\subsubsection{Quantization Decision}
\label{subsubsec:configuration-decision}

\SYS{} makes the same decision with QServe~\cite{qserve} on activation quantization to use 8-bit, as the 4-bit activation can lead to a large accuracy drop but does not lead to significant system efficiency improvement as MatMul execution tends to be bound more on the larger weight tensor rather than the smaller activation tensor according to the compute intensity $I = \frac{2MNK}{MKB_{act} + KNB{weight}}$ ($M$ is the token number, $K$/$N$ are the in/out features, and $B_{act}$ and $B_{weight}$ are the bytes per element of activation and weight).

Instead of using per-token activation quantization, \SYS{} uses group-wise RTN method.
On the one hand, Tab.\ref{tab:ppl-all} shows that simple group-wise RTN quantization outperforms token-wise smoothing method.
On the other hand, the weight is already group-wise in \SYS{}, and the group-wise activation does not lead to significantly more dequantization overhead in the system.
We observe symmetric quantization is enough for the 8-bit activation (refer to \SYS{} W8A8 in Tab.\ref{tab:ppl-all}), while asymmetric is essential for the 4-bit weight.
The group-wise method with asymmetric can lead to difficulty for the kernel to make use int8 Tensor Core, for which we design a \textit{two-step dequantization} method with the property of the mix of symmetric and asymmetric (Sec.\ref{subsec:system}).

\subsection{Global Precision Search Algorithm}
\label{subsec:search}

\SYS{} determines the precision of all output features in all layers globally.
It identifies the salience of these features with respect to the final loss of the model, and assigns larger bit-width to the features leading to larger loss.
Specifically, it calculates the salience $S$ of a channel $c$ as:
\begin{equation}
S_c = \lvert l(c_q) - l(c_0) \rvert
\label{eq:salience}
\end{equation}
which is the distance of the model's loss between quantizing and not quantizing channel $c$.
In Eq.\ref{eq:salience}, $l()$ is the loss function of the model w.r.t. a single channel, $c_q$ is the quantized weight of the channel and $c_0$ is the original one.

We use the Taylor Expansion method to estimate the loss function $l(c)$, ignoring the high-order items:
\begin{equation}
l(c) \approx l(c_0) + g^T (c-c_0) + \frac{1}{2}(c-c_0)^T H (c-c_0)
\end{equation}
where $g = \mathbb E[\frac{\partial}{\partial c} l(c)]$ is the loss's gradient w.r.t. the channel, and $H = \mathbb E[\frac{\partial^2}{\partial c^2} l(c)]$ is the second-order gradient (i.e., Hessian matrix).
It is infeasible to calculate the Hessian matrix.
We approximate the Hessian with the (empirical) Fisher information matrix (FIM) $F$ on the calibration dataset $\mathcal{D}$:
\begin{equation}
H \approx F = \frac{1}{\lvert \mathcal{D} \rvert} \sum_{d \in \mathcal{D}}{g_d g_d^T}
\end{equation}
Note that $F$ is w.r.t. a channel,
differing from the diagonal FIM in the recent works that ignores any cross-neuron interactions~\cite{transformer-prune-2022, squeezellm}.

Based on this approximation, the second order loss factor $\frac{1}{2} (c-c_0)^T (g_d g_d^T) (c-c_0)$ can be further simplified to $ \frac{1}{2}(g_d^T (c-c_0))^2$, 
simplifying the expensive chained matrix multiplication into a single vector product. Finally, the salience can be calculated by:
\begin{equation}
S_c = \frac{1}{\lvert \mathit{D} \rvert} \sum_{d \in \mathit{D}} \lvert g_d^T(c_q - c_0) + \frac{1}{2}(g_d^T(c_q - c_0))^2 \rvert
\label{eq:salience_final}
\end{equation}

We do not ignore the first order information during the calculation, differing from the recent quantization works~\cite{gptq,spqr,squeezellm}.
Note that what we require is the loss itself rather than the arguments of the loss function,
thus we do not need to ignore the first order factor to simplify the arguments calculation.

\begin{algorithm}[ht]
\caption{Global precision search procedure.}
\label{alg:precision-search}
\begin{algorithmic}[1]
\REQUIRE Weight and gradient of all linear layers ($W_i \in \mathbb{R}^{O \times I}$, $G_i \in \mathbb{R}^{O \times I}$ for layer $i \in [1..L]$). Total number of output channels with large bit-width ($T$).
\ENSURE Channel index for large and small bit-width ($\mathcal{C}^{lg}$ and $\mathcal{C}^{sm}$).
\STATE $\mathcal{C}^{global}$ $\leftarrow$ ()
\COMMENT {\cgray{Global channel information}}
\FOR {$i=1$ {\bfseries to} $L$}
\STATE $W_i^{delta}$ $\leftarrow$ quantize($W_i$) - $W_i$
\STATE $S^{1st}$ $\leftarrow$ sum($G_i \odot W_i^{delta}$, dim=1)
\STATE $S^{2nd} \leftarrow 0.5 * (S^{1st})^2$
\STATE $S \leftarrow \lvert S^{1st} + S^{2nd} \rvert$
\COMMENT {\cgray{$O$ channels' salience $\in \mathbb{R}^{O}$}}
\FOR {$cid = 1$ {\bfseries to} $O$}
\STATE $\mathcal{C}^{global}$ $\leftarrow$ $\mathcal{C}^{global}$ $\cup$ ( tuple($i$, $cid$, $S_{cid}$) )
\ENDFOR
\ENDFOR
\STATE Sort $\mathcal{C}^{global}$ by salience in descending order.
\STATE $\mathcal{C}^{lg}, \mathcal{C}^{sm} \leftarrow \mathcal{C}^{global}_{:T}, \mathcal{C}^{global}_{T:}$
\end{algorithmic}
\end{algorithm}

Algo.\ref{alg:precision-search} illustrates the procedure of the global precision search.
It calculates the salience of all the output channels of all linear layers and sort them in descending order.
Given the global threshold $T$ as the number of large-bit precision channels, the first $T$ channels are intended to be quantized with 8-bit, and the other channels will be quantized with smaller bit-width (i.e., 4-bit in this paper).

\subsection{Efficient Quantization Computation System}
\label{subsec:system}

\begin{figure*}[!ht]
    \centering
    \includegraphics[width=.95\textwidth]{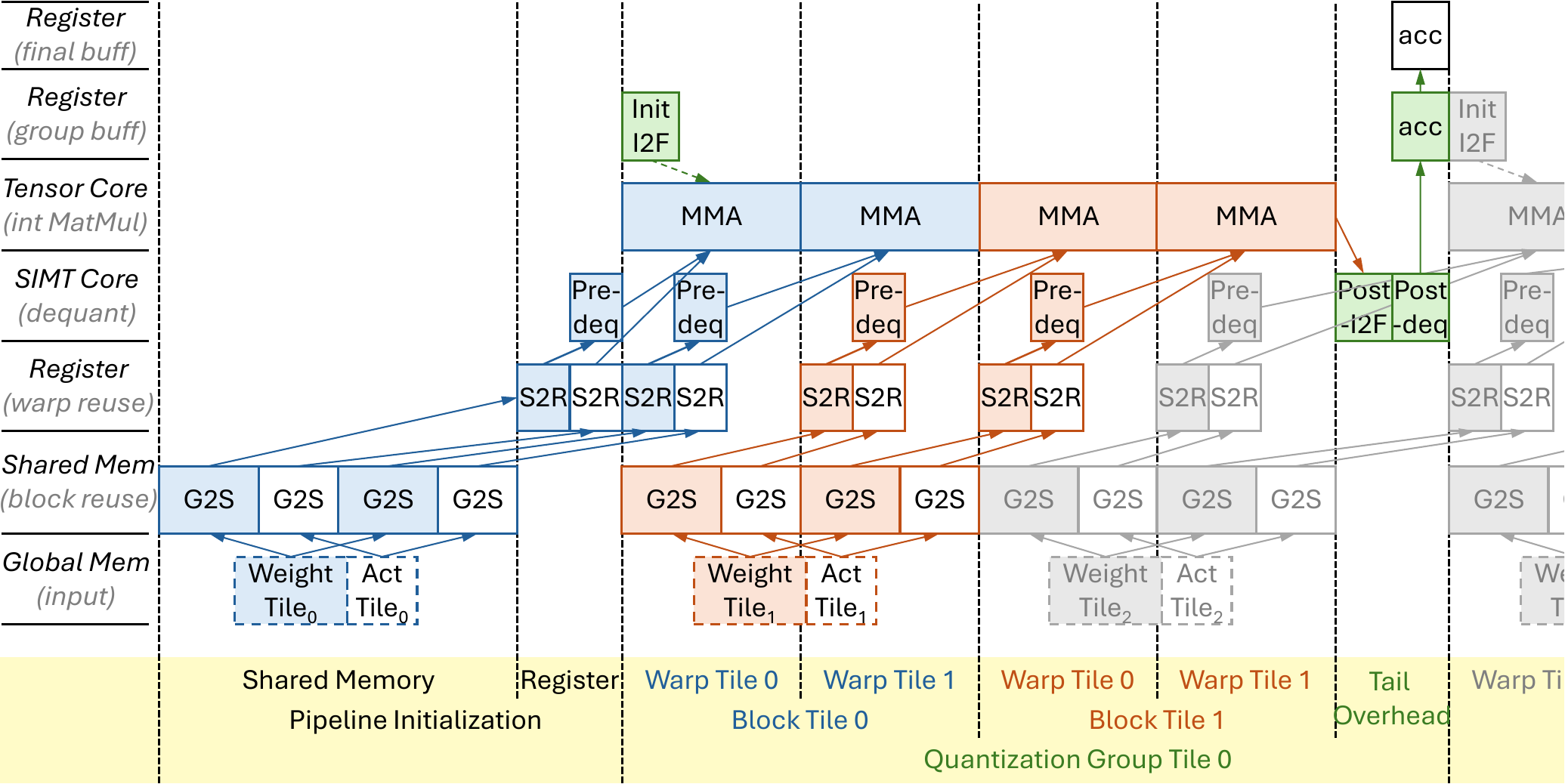}
    \caption{The GPU kernel software pipeline of group-wise W4A8/W8A8 quantized MatMul. 
    It assumes perfect overlapping.
    G2S: load global to shared memory;
    S2R: load shared memory to register; 
    MMA: matrix multiply-accumulation;
    I2F: integer to float conversion;
    deq: dequantize;
    acc: accumulate.
    \cb{While this pipeline is modeled on the NVIDIA A100 architecture, its fundamental principles remain applicable to subsequent generations, such as Hopper and Blackwell, subject to minor architectural refinements. For instance, newer architectures can utilize the Tensor Memory Accelerator (TMA) to load activation tensors directly, bypassing registers before they reach the Tensor Cores. Furthermore, these architectures support warp specialization for memory loading as an alternative to a uniform execution scheme.
    }
    }
    \label{fig:kernel-software-pipeline}
\end{figure*}

\textbf{Two-step dequantization to make use of int8 Tensor Core.}
As for the W4A8 computation, the dequantized weight and activations are $(W_q - z)s_w$ and $A_q s_a$, where $W_q$ and $z$ are uint4 datatype (4-bit unsigned integer), $A_q$ is int8 datatype, and $s_w$ and $s_a$ are float16 datatype.
Directly dequantizing the tensors into float16 datatype before the MatMul computation will prevent us using the fast 8-bit Tensor Core on the GPU.
Instead, \SYS{} uses a two step dequantization within each group.
Specifically, \SYS{} first partially dequantizes the weight into $(W_q - z)$, and then multiply it by $A_q$ with the 8-bit Tensor Core.
Finally, it multiplies this MatMul result by the two scales within each group.
Note that we use int8 datatype for $(W_q - z)$ so that there is no overflow problem.

\begin{figure}[ht]
    \centering
    \includegraphics[width=0.85\columnwidth]{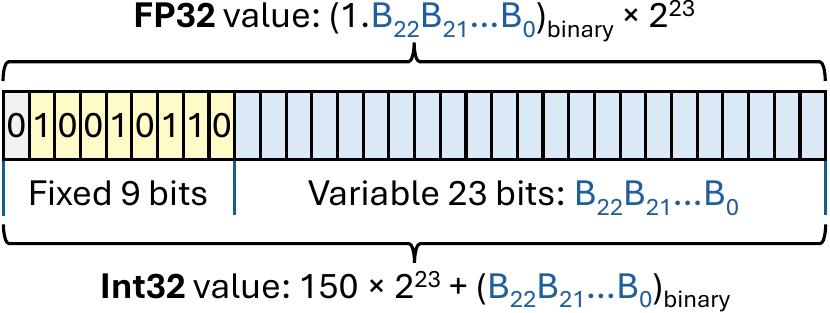}
    \caption{The float and integer value of binary \texttt{(010010110xx...x)}, each within a consecutive range.}
    \label{fig:int-float-representation}
\end{figure}

\textbf{Fast I2F with partially fusing into Tensor Core instruction.}
In the above two-step dequantization computation, the step 2 is the multiplication between integer and float tensor, requiring integer to float conversion (I2F).
As I2F instruction is expensive on the modern GPUs, we make use of the range-dependent fast I2F transformation to convert the I2F instruction into two add/sub instructions.
Specifically, it is based on the fact that there exists a certain range where an integer value's binary is the same as a corresponding float binary.
As shown in Fig.\ref{fig:int-float-representation}, the binary with the first 9 bits as \texttt{010010110} represents a series of consecutive int32 and float32 values, respectively.
We can add a bias to an integer value to make it within this consecutive range, and then subtract a corresponding bias in float to restore its value in float.
We take the middle value in this range as the bias to maximize the data range that can be safely converted, whose hexadecimal number is $mid = \texttt{0x4b400000}$ (i.e., in the remaining 23 bits, the first bit is 1 and the other bits are 0).
This allows to convert a consecutive range of $2^{23}$ int32 numbers to float32.
The range of dot product of $k$ int8 elements is in $2^{16}k$, thus the above fast I2F conversion allows the $k$ value to be 128.
We use quantization group size as 128 and can use the fast I2F safely:

\begin{lstlisting}[language=C, basicstyle=\footnotesize, numbers=left, xleftmargin=0.2in, breaklines=true, commentstyle=\color{gray}\textit, frame=tb, columns=fullflexible]
// bias_int = as_int(mid), bias_fp = as_float(mid);
int tmp = src_int + bias_int;
int dst_float = *((float*)&tmp) - bias_fp;
\end{lstlisting}

We further fuse the integer subtraction into the Tensor Core mma (Matrix Multiply-Accumulate $D = AB + D$) instruction.
We initialize the accumulator $D$ as the \texttt{bias\_int} before MatMul computation of each quantization group, and will only need to subtract the \texttt{bias\_fp} after the MatMul.
In another word, the expensive I2F is converted into a single float subtraction.
The above I2F simplification brings more than 20 TOPS performance improvement for the 512/4096/4096 (M/N/K) quantized MatMul computation on an A100 GPU.

\textbf{End-to-end software pipeline of the quantized linear kernel on the GPU.}
Fig.\ref{fig:kernel-software-pipeline} shows the software pipeline of the quantized kernel.
Besides the basic warp tile and block tile, we introduce the quantization group tile for the per-group quantized computation.
It uses two output buffers for the output accumulation at register level, one for the per-group accumulation, and the other for the global accumulation.
This allows to apply the per-group scales on the group-level buffer.
We initialize the group buffer with the \texttt{bias\_int} at the beginning of the group tile, and subtract \texttt{bias\_fp} at the end of the group tile.
As for the \textit{two-step dequantization}, the first step is within the warp tile where each input element will subtract the zero-point before feeding into MMA, the second step is at the end of the group tile by multiplying the scale.
We use the vectorized intrinsic to perform four int8 subtract in a single instruction (\texttt{vsub4})~\cite{qserve}.
Besides, to improve the global memory loading efficiency, we prepack the memory layout of the weight tensor ahead-of-time to avoid the runtime permutation of the input elements.
This software pipeline can overlap the memory loading, the dequantization computation with SIMT Core, and the MatMul computation with Tensor Core to the best, and minimizes the overhead of group-wise dequantization.

\subsection{Parallel Execution of Sub-problems of Different Bit-width}

As for the execution shown in Fig.\ref{fig:mixed-precision-linear},
\SYS{} executes different sub-problems in parallel on the GPU with CUDA Graph.
Finally, the MatMul execution of the two parts write to the same target tensor with different channel indices to generate the final output.
We implement this function with the fused epilogue of MatMul to scatter the output to the corresponding indices, which is basically costless.

\subsection{\cb{Quantization-aware Roofline Discussion}}

\cb{
To systematically analyze the performance bounds of \SYS{} quantization schemes, we extend the traditional Roofline model to account for the heterogeneous compute pipelines on NVIDIA GPUs.
Using the NVIDIA A100 (80GB SXM) as our target architecture, we evaluate the system against its peak hardware limits: 2039 GB/s of High Bandwidth Memory (HBM2e) throughput, 624 TOPS of dense INT8 Tensor Core compute, and 19.5 TFLOPS of FP32 SIMT compute.
Introducing sub-byte quantization and asymmetric zero-points fundamentally alters the arithmetic intensity (operations per byte) and introduces multiple competing compute ceilings across the execution units.
}

\cb{
\textbf{Memory Traffic.}
From a memory bandwidth perspective, arithmetic intensity is governed by the quantized data footprint and associated metadata. For a group size of $G=128$, the W8A8 symmetric scheme fetches 8-bit weights, 8-bit activations, and a 16-bit scale, yielding an effective weight footprint of $1 + 2/G$ bytes per element.
The W4A8 asymmetric scheme compresses weights to 4 bits, requiring a 16-bit scale and a 4-bit zero-point, yielding an effective footprint of $0.5 + 2.5/G$ bytes per weight.
Consequently, W4A8 drastically shifts the memory-bound slope upward, allowing significantly higher attainable throughput in bandwidth-constrained regimes, such as small batch decoding.
This motivates quantization with lower bits for weight in \SYS{}.
}

\cb{
\textbf{Compute Ceilings: Tensor Core vs. SIMT Dequantization.}
In the compute-bound regime, performance is dictated by the interplay between the Tensor Cores (handling the GEMM) and the SIMT cores (handling dequantization). 
}

\cb{
For both W8A8 and W4A8, output scaling requires a transition from the Tensor Core's \texttt{int32} accumulators to the SIMT core's \texttt{float32} pipeline.
For every group size of $G=128$, the Tensor Cores perform 128 INT8 multiply-accumulate (MAC) operations (256 operations total) to produce a single \texttt{int32} partial sum.
Before the FP32 scales ($s_w$ and $s_a$) can be applied, the SIMT cores must first issue an \texttt{int32}-to-\texttt{float32} cast instruction.
Therefore, the post-GEMM workload per group consists of this explicit cast instruction followed by the FP32 scale multiplications.
Even with this added casting overhead, the application's required operational ratio—256 INT8 operations per 3 SIMT instructions (cast + multiplies)—remains well below the A100 hardware's provisioned ratio of 32:1 (624 INT8 TOPS vs. 19.5 FP32 TFLOPS), thanks to the fast I2F and partial fusion methodology in \SYS{}.
Because the SIMT requirement is so light, this post-GEMM casting and scaling can be hidden behind the Tensor Core accumulation.
This further motivates the group-wise 8-bit quantization than per-channel/per-token solutions.
}

\cb{
However, W4A8 introduces a heavier pre-GEMM dequantization step: $$Y = ((W_q - z) \times A_q) \cdot s_w \cdot s_a$$
Before the Tensor Cores can execute the INT8 MACs, the 4-bit weights must be unpacked and the zero-point subtracted.
By leveraging the \texttt{vsub4} intrinsic, the hardware can perform 4 \texttt{int8} subtractions within a single SIMT instruction, highly optimizing the $(W_q - z)$ computation.
Despite this instruction-level parallelism, the SIMT integer pipeline must still continuously feed the Tensor Cores.
Because this pre-GEMM workload scales at $O(N^2)$ and competes for instruction fetch and register bandwidth concurrently with the $O(N^3)$ Tensor Core math, it creates an auxiliary, lower compute ceiling.
Thus, the effective peak compute for W4A8 is strictly lower than the absolute 624 TOPS achieved by W8A8.
}

\cb{
\textbf{Intersection Point (Ridge Point) Analysis.}
The traditional Roofline ridge point defines the intersection between memory bandwidth and peak compute, marking the transition from a memory-bound to a compute-bound workload.
For pure INT8 Tensor Core execution on the A100, this intersection occurs at roughly 306 OPs/Byte (624 TOPS / 2039 GB/s).
Because the W4A8 scheme is constrained by the SIMT \texttt{vsub4} dequantization ceiling rather than the peak Tensor Core ceiling, its horizontal Roofline ceiling drops.
Consequently, the intersection between the W4A8 memory traffic line and its compute ceiling shifts to the left.
This leftward shift indicates that W4A8 workloads saturate their maximum achievable compute at a lower arithmetic intensity than W8A8, entering the compute-bound regime earlier due to SIMT instruction overhead.
}


\section{Evaluation}
\label{sec:evaluation}

\begin{table*}[ht]
\center
\scriptsize
\caption{Perplexity evaluation ($\downarrow$) on wikitext2/c4 (gray for c4), sequence length 2048.
\texttt{NA} means no support. 
\texttt{Abn} means the value is too large ($>10^5$).
For \SYS{}, \texttt{pn} means $n\%$ 8-bit.
}
\label{tab:ppl-all}
\begin{tabular}{@{}cc|ccc|cccc|c@{}}
\toprule
\multicolumn{2}{c|}{} &
  \multicolumn{3}{c|}{Llama 3.1/3.2} &
  \multicolumn{4}{c|}{Qwen2.5} &
  Mistral \\
\multicolumn{2}{c|}{\multirow{-2}{*}{baselines}} &
  1B &
  8B &
  70B &
  0.5B &
  1.5B &
  7B &
  32B &
  7B v0.3 \\ \midrule
\multicolumn{2}{c|}{float16} &
  9.75/\cgray{12.72} &
  6.24/\cgray{8.95} &
  2.81/\cgray{6.68} &
  13.07/\cgray{17.55} &
  9.26/\cgray{13.11} &
  6.85/\cgray{10.44} &
  5.02/\cgray{8.95} &
  5.32/\cgray{7.84} \\ \midrule
 &
  W4A16 &
  11.72/\cgray{15.56} &
  6.82/\cgray{9.72} &
  3.55/\cgray{7.43} &
  15.54/\cgray{20.55} &
  10.35/\cgray{14.35} &
  7.23/\cgray{10.88} &
  5.27/\cgray{9.14} &
  5.51/\cgray{8.04} \\
\multirow{-2}{*}{Basic RTN} &
  W5A16 &
  10.15/\cgray{13.25} &
  6.40/\cgray{9.15} &
  3.16/\cgray{9.52} &
  13.61/\cgray{18.17} &
  9.52/\cgray{13.38} &
  6.95/\cgray{10.53} &
  5.09/\cgray{8.99} &
  5.38/\cgray{7.91} \\ \midrule
SmoothQuant &
   W8A8 &
   9.89/\cgray{12.91} &
   6.34/\cgray{9.08} &
   2.92/\cgray{6.77} &
   13.84/\cgray{18.40} &
   9.63/\cgray{13.49} &
   7.17/\cgray{10.85} &
   5.12/\cgray{9.04} &
   5.35/\cgray{7.88} \\
 &
  W4A4 &
  Abn/\cgray{Abn} &
  8.34/\cgray{11.95} &
  6.16/\cgray{9.91} &
  NA/\cgray{NA} &
  Abn/\cgray{Abn} &
  8.15/\cgray{12.05} &
  6.26/\cgray{9.98} &
  5.83/\cgray{8.50} \\
\multirow{-2}{*}{QuaRot} &
  W4A8 &
  Abn/\cgray{Abn} &
  6.60/\cgray{9.67} &
  3.43/\cgray{7.10} &
  NA/\cgray{NA} &
  Abn/\cgray{Abn} &
  7.03/\cgray{10.68} &
  5.23/\cgray{9.10} &
  5.40/\cgray{7.99} \\ 
QServe &
  W4A8 &
  Abn/\cgray{Abn} &
  6.64/\cgray{9.49} &
  3.49/\cgray{7.07} &
  Abn/\cgray{Abn} &
  Abn/\cgray{Abn} &
  7.39/\cgray{11.06} &
  5.55/\cgray{9.31} &
  5.44/\cgray{7.98} \\ \midrule
 &
  W4A8 (p0) &
  10.36/\cgray{14.09} &
  6.54/\cgray{9.62} &
  3.30/\cgray{7.24} &
  14.43/\cgray{19.61} &
  9.66/\cgray{13.79} &
  7.03/\cgray{10.75} &
  5.21/\cgray{9.08} &
  5.42/\cgray{8.02} \\
 &
  W4.4A8 (p10) &
  10.05/\cgray{13.51} &
  6.42/\cgray{9.33} &
  3.02/\cgray{6.83} &
  13.42/\cgray{18.13} &
  9.44/\cgray{13.43} &
  6.92/\cgray{10.57} &
  5.12/\cgray{9.01} &
  5.36/\cgray{7.93} \\
 &
  W4.8A8 (p20) &
  9.95/\cgray{13.25} &
  6.37/\cgray{9.22} &
  2.97/\cgray{6.79} &
  13.32/\cgray{17.99} &
  9.40/\cgray{13.35} &
  6.90/\cgray{10.53} &
  5.09/\cgray{9.00} &
  5.35/\cgray{7.90} \\
 &
  W6A8 (p50) &
  9.85/\cgray{12.98} &
  6.30/\cgray{9.09} &
  2.86/\cgray{6.73} &
  13.21/\cgray{17.78} &
  9.33/\cgray{13.25} &
  6.88/\cgray{10.49} &
  5.05/\cgray{8.98} &
  5.33/\cgray{7.87} \\
\multirow{-5}{*}{\SYS{}} &
   W8A8 (p100) &
   9.76/\cgray{12.75} &
   6.25/\cgray{8.97} &
   2.81/\cgray{6.68} &
   13.12/\cgray{17.60} &
   9.28/\cgray{13.14} &
   6.86/\cgray{10.45} &
   5.02/\cgray{8.96} &
   5.32/\cgray{7.84} \\ \bottomrule
\end{tabular}
\end{table*}

\subsection{Setup}
\label{subsec:setup}

As for \SYS{} evaluation, we use 0\%, 10\%, 20\%, 50\% and 100\% of 8-bit based on the 4-bit quantization, respectively.
Meanwhile, we use 8-bit for activation quantization.
Both the weight and activation are group-wise quantized with group size 128.
The 4-bit part is asymmetric quantized and the 8-bit part (including that in weight) is symmetric.
Similar to the recent work~\cite{qserve, spinquant}, we enable GPTQ and clipping in \SYS{}.

\textbf{Baselines and configurations.}
In Sec.\ref{subsec:ppl-eval} and Sec.\ref{subsec:downstream-eval}, we compare to the pure PTQ methods of weight-activation quantization, neither the QAT nor the methods training the quantization parameters (e.g., training the rotation matrix like in SpinQuant\cite{spinquant}).
Specifically, we compare \SYS{} with the SOTA PTQ weight-activation methods, including the most widely used SmoothQuant~\cite{smoothquant} and the recent SOTA QuaRot\cite{quarot} (of both W4A4 and W4A8) and QServe~\cite{qserve}.
The 8-bit tensors are all symmetric quantized in all baselines.
We use symmetric and per-channel/token quantization in QuaRot, following the setting in its paper.
We follow the official configurations to use 0.85/0.15 as the alpha/beta parameter for SmoothQuant, and 0.3/0.7 for QServe.
We disable the KV quantization of QuaRot and QServe in our experiments to make the comparison fair.
In Appendix.\ref{subsec:compare-more-related}, We also compare the ppl with GPTQ~\cite{gptq}, AWQ\cite{awq}, SqueezeLLM\cite{squeezellm}, OmniQuant\cite{OmniQuant}, AffineQuant\cite{affinequant}, Atom\cite{atom} and SpinQuant\cite{spinquant}.


\textbf{Models and Datasets.}
We evaluate \SYS{} and the baselines on Llama 3.1 8B and 70B~\cite{llama3}, Llama 3.2 1B, Qwen2.5 0.5B, 1.5B, 7B and 32B~\cite{qwen2.5}, and Mistral 7B v0.3~\cite{mistral7b}.
We use wikitext2~\cite{wikitext2} as the calibration set for \SYS{}, 
and the default pile dataset~\cite{pileval} for SmoothQuant and QServe.
\SYS{} uses 128 samples with sequence length of 2048.
SmoothQuant and QServe uses 64 samples with sequence length of 1024 to prevent OOM.

\textbf{Metrics.}
We compare the perplexity (ppl) between all the baselines on wikitext2 and C4~\cite{c4} datasets.
We also compare a set of popular downstream tasks on Llama 3.1 8B, Qwen2.5 7B, and Mixtral 7B v0.3 through lm-eval~\cite{eval-harness}, 
including BBH~\cite{bbh}, GPQA~\cite{gpqa}, MMLU-Pro~\cite{mmlu-pro}, MuSR~\cite{musr}, ARC challenge~\cite{arc}, and HellaSwag~\cite{hellaswag}.
We conduct the system experiments on NVIDIA A100 (80G) GPUs with CUDA 12.1, \cb{except for that Sec.\ref{subsubsec:perf-h100} discusses the performance on NVIDIA H100 GPU}.
We use PyTorch 2.4.1 and transformers 4.45.2.

\subsection{Perplexity Evaluation}
\label{subsec:ppl-eval}

Tab.\ref{tab:ppl-all} shows the perplexity on Wikitext2 and C4 dataset for the commonly used open source LLMs, of different baselines.
It shows that:
\textbf{1)} Using 4.4 bits of weights with \SYS{} can achieve the similar accuracy with 5 bits RTN weight-only quantization, even with 8-bit activation quantization enabled in \SYS{}.
This is mainly because \SYS{} assigns the high-salience output channels with larger bit-widths than the uniform 5-bit solution.
\textbf{2)} As for the weight-activation quantization baselines, \SYS{} W4.4A8 shows a comparable accuracy with SmoothQuant with much smaller bit-width (60\% of that in SmoothQuant).
\SYS{} W4.4A8 shows better accuracy than QuaRot and QServe with only 10\% larger bit-width.
It shows \SYS{} achieves a good balance of memory consumption and accuracy.
\textbf{3)} Note that \SYS{} W8A8 quantization shows nearly lossless performance compared to the float16 baseline.
Besides, the \SYS{} W4A8 also outperforms the SOTA QuaRot and QServe for many cases, due to using group-wise quantization for the activation in \SYS{} rather than the per-token method in QuaRot and QServe.
This is part of the motivation that \SYS{} uses group-wise quantization for the activation.

\begin{figure*}[ht]
    \centering
    \includegraphics[width=0.85\textwidth]{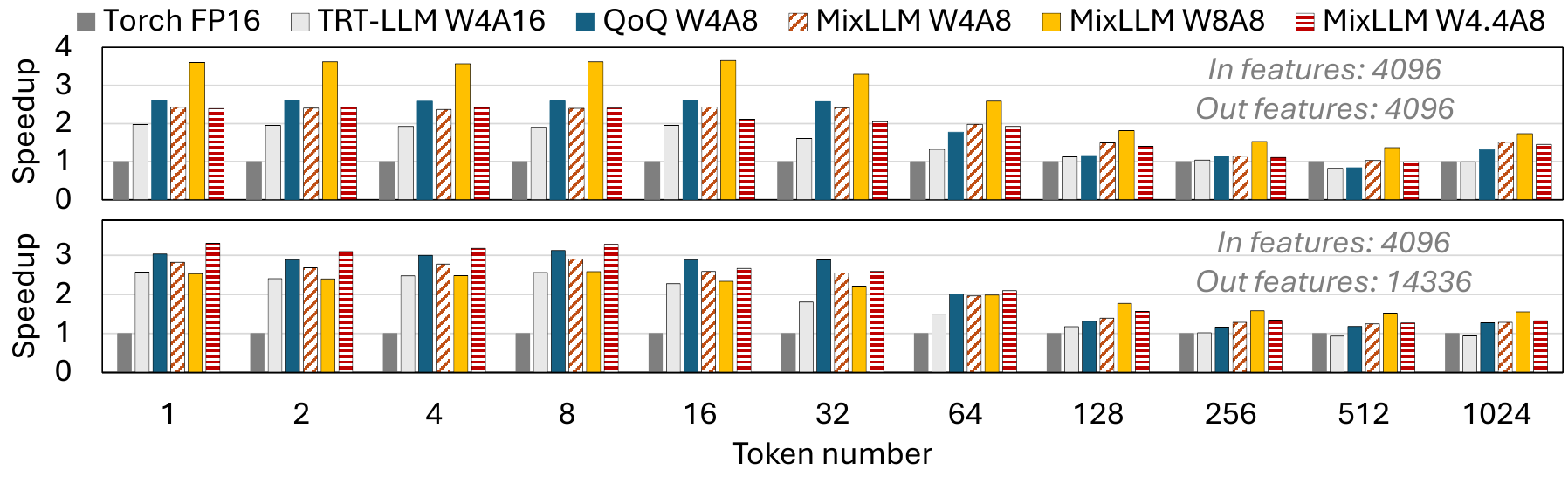}
    \caption{The speedup of two types of single linear layers over torch FP16 baseline on A100 GPU.}
    \label{fig:linear-speedup}
\end{figure*}

\subsection{System Performance}

\subsubsection{Performance on A100 GPU}

We have evaluated \SYS{} for the single linear layer of token numbers ranging from 1 to 1024 with in features 4096 and out features 4096/14336, and compared it with the SOTA W4A16 (TRT-LLM) and QServe~\cite{qserve}, shown in Fig.\ref{fig:linear-speedup}.
It also shows \SYS{} kernel performance of different percent of 8-bits (W4A8 0\% 8-bit, W4.4A8 10\% 8-bit, and W8A8 100\% 8-bit).
It shows that:
\textbf{1)} \SYS{} outperforms the float16 counterpart for all token numbers, achieving 1.96$\times$, 2.76$\times$, and 1.88$\times$ averaged speedup with \SYS{} W4A8, W8A8, and W4.8A8 respectively for output feature 4096, and 2.45$\times$, 2.15$\times$, 2.34$\times$ for output feature 14336.
\textbf{2)} \SYS{} outperforms the SOTA W4A16 solution, achieving 1.31$\times$, 1.80$\times$, and 1.25$\times$ averaged speedup with \SYS{} W4A8, W8A8, and W4.8A8 respectively for output feature 4096, and 1.36$\times$, 1.31$\times$, 1.33$\times$ for output feature 14336.
\textbf{3)} \SYS{} achieves better performance than QServe with similar bit-width, achieving 1.03$\times$, 1.41$\times$, and 0.99$\times$ averaged speedup with \SYS{} W4A8, W8A8, and W4.8A8 respectively for output feature 4096, and 1.08$\times$, 1.04$\times$, 1.05$\times$ for output feature 14336.

\begin{figure}[]
    \centering
    \includegraphics[width=\columnwidth]{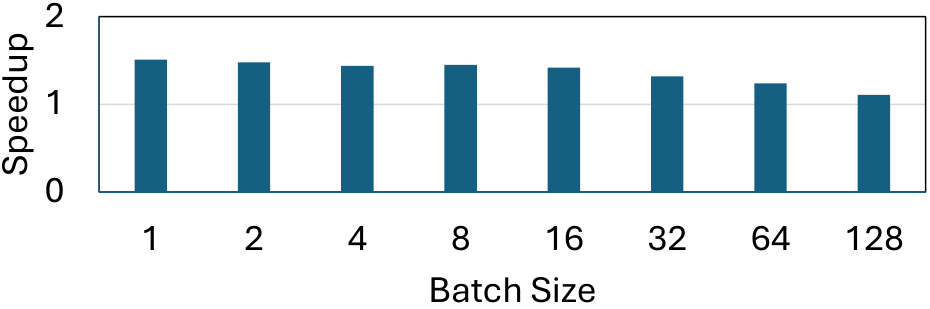}
    \caption{Speedup of \SYS{} W8A8 of Qwen 2.5 7B over float16 baseline on A100 GPU.}
    \label{fig:w8a8-batch-perf}
\end{figure}

We have also integrated \SYS{} into vLLM, achieving 1.41$\times$ speedup of output token throughput (token/sec) than the float16 baseline given the configuration of batchsize 2 and input/output length 1000/1000, using W4.4A8 for Mixtral-7B on a single A100 GPU.
\cb{
Fig.\ref{fig:w8a8-batch-perf} shows the speedup over float16 baseline using \SYS{} W8A8 quantization for Qwen2.5 7B model on A100 GPU, with input/output length 1000/1000 and a range of batch size from 1 to 128.
}

\subsubsection{\cb{Performance on H100 GPU}}
\label{subsubsec:perf-h100}

\begin{figure}[]
    \centering
    \includegraphics[width=1.0\columnwidth]{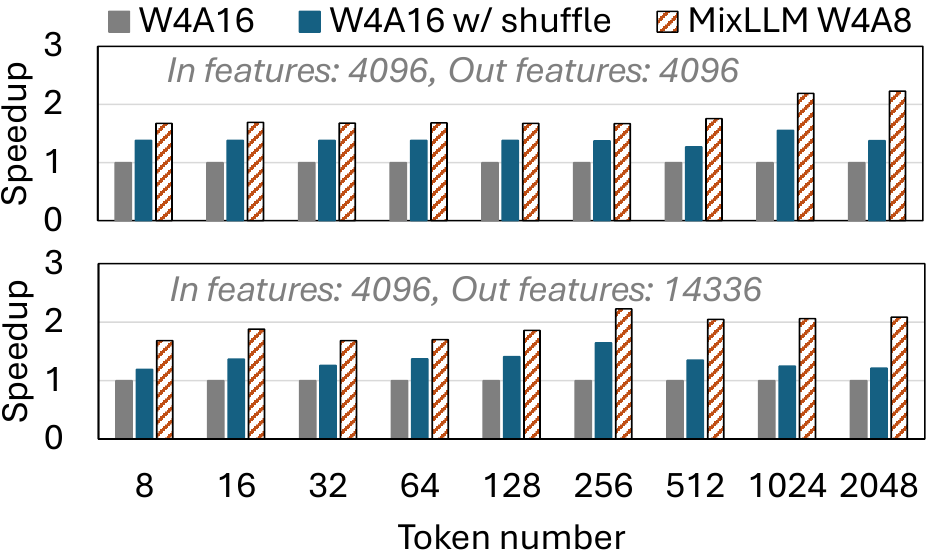}
    \caption{\cb{The speedup of two types of single linear layers over W4A16 baseline on H100 GPU.}}
    \label{fig:linear-speedup-h100}
\end{figure}

\cb{
We evaluate the performance of single linear layers on the NVIDIA H100 GPU, comparing the \SYS{} W4A8 kernel against the W4A16 baseline (Fig. \ref{fig:linear-speedup-h100}).
To the best of our knowledge, an H100-optimized implementation of the QoQ solution~\cite{qserve} is currently unavailable.
The W4A16 baseline was established by identifying the optimal configuration through exhaustive tuning of the official CUTLASS group-wise and asymmetric-quantized W4A16 examples.
Our analysis considers both a naive 4-bit layout and an advanced shuffled 4-bit layout.
Results indicate that \SYS{} consistently outperforms the W4A16 solutions across all configurations.
Specifically, for an output feature size of 4096, \SYS{} achieves average speedups of $1.81\times$ and $1.39\times$ over the naive and shuffled layouts, respectively; for a size of 14336, these speedups are $1.91\times$ and $1.34\times$.
}

\begin{table*}[ht]
\center
\scriptsize
\caption{Downstream tasks evaluation ($\uparrow$) on Llama-3.1-8B/Qwen2.5-7B/Mistral-7B-v0.3.
The above is the average of the three models.
BBH is 3 shot, MMLU pro is 5 shot, and others are zero shot.}
\label{tab:zero-shot}
\begin{tabular}{@{}ccccccc@{}}
\toprule
 &
  BBH &
  GPQA &
  MMLU-Pro &
  MuSR &
  ARCc &
  HellaSwag \\ \midrule
float16 &
  \begin{tabular}[c]{@{}c@{}}48.62\\ \cgray{$^{46.52/54.09/45.25}$}\end{tabular} &
  \begin{tabular}[c]{@{}c@{}}30.86\\ \cgray{$^{31.08/33.11/28.39}$}\end{tabular} &
  \begin{tabular}[c]{@{}c@{}}35.52\\ \cgray{$^{32.91/43.86/29.80}$}\end{tabular} &
  \begin{tabular}[c]{@{}c@{}}41.07\\ \cgray{$^{37.99/44.51/40.72}$}\end{tabular} &
  \begin{tabular}[c]{@{}c@{}}52.24\\ \cgray{$^{53.41/51.02/52.30}$}\end{tabular} &
  \begin{tabular}[c]{@{}c@{}}79.43\\ \cgray{$^{78.92/78.94/80.43}$}\end{tabular} \\ \midrule

SmoothQuant W8A8 &
  \begin{tabular}[c]{@{}c@{}}47.82\\ \cgray{$^{46.37/52.57/44.52}$}\end{tabular} &
  \begin{tabular}[c]{@{}c@{}}30.90\\ \cgray{$^{31.40/33.94/27.36}$}\end{tabular} &
  \begin{tabular}[c]{@{}c@{}}35.04\\ \cgray{$^{32.61/42.98/29.52}$}\end{tabular} &
  \begin{tabular}[c]{@{}c@{}}42.06\\ \cgray{$^{39.05/46.39/40.73}$}\end{tabular} &
  \begin{tabular}[c]{@{}c@{}}51.74\\ \cgray{$^{53.33/50.00/51.88}$}\end{tabular} &
  \begin{tabular}[c]{@{}c@{}}79.20\\ \cgray{$^{78.88/78.48/80.24}$}\end{tabular} \\
QuaRot W4A4 &
  \begin{tabular}[c]{@{}c@{}}41.10\\ \cgray{$^{36.96/45.42/40.92}$}\end{tabular} &
  \begin{tabular}[c]{@{}c@{}}27.53\\ \cgray{$^{25.41/28.94/28.23}$}\end{tabular} &
  \begin{tabular}[c]{@{}c@{}}27.60\\ \cgray{$^{22.99/34.40/25.42}$}\end{tabular} &
  \begin{tabular}[c]{@{}c@{}}39.46\\ \cgray{$^{37.92/40.68/39.77}$}\end{tabular} &
  \begin{tabular}[c]{@{}c@{}}45.99\\ \cgray{$^{43.00/46.33/48.63}$}\end{tabular} &
  \begin{tabular}[c]{@{}c@{}}74.85\\ \cgray{$^{72.87/73.54/78.14}$}\end{tabular} \\

QuaRot W4A8 &
  \begin{tabular}[c]{@{}c@{}}46.95\\ \cgray{$^{44.95/52.98/42.92}$}\end{tabular} &
  \begin{tabular}[c]{@{}c@{}}30.28\\ \cgray{$^{30.96/30.71/29.18}$}\end{tabular} &
  \begin{tabular}[c]{@{}c@{}}33.60\\ \cgray{$^{29.95/42.45/28.41}$}\end{tabular} &
  \begin{tabular}[c]{@{}c@{}}41.65\\ \cgray{$^{39.05/45.58/40.32}$}\end{tabular} &
  \begin{tabular}[c]{@{}c@{}}51.39\\ \cgray{$^{50.00/52.30/51.88}$}\end{tabular} &
  \begin{tabular}[c]{@{}c@{}}78.55\\ \cgray{$^{77.83/77.84/79.98}$}\end{tabular} \\ 

QServe W4A8 &
  \begin{tabular}[c]{@{}c@{}}45.78\\ \cgray{$^{40.98/51.23/45.14}$}\end{tabular} &
  \begin{tabular}[c]{@{}c@{}}30.02\\ \cgray{$^{28.99/32.50/28.56}$}\end{tabular} &
  \begin{tabular}[c]{@{}c@{}}32.84\\ \cgray{$^{28.16/41.72/28.63}$}\end{tabular} &
  \begin{tabular}[c]{@{}c@{}}39.92\\ \cgray{$^{37.60/41.59/40.57}$}\end{tabular} &
  \begin{tabular}[c]{@{}c@{}}50.54\\ \cgray{$^{51.28/49.15/51.19}$}\end{tabular} &
  \begin{tabular}[c]{@{}c@{}}78.10\\ \cgray{$^{76.90/77.52/79.89}$}\end{tabular} \\ \midrule
\SYS{} W4A8 &
  \begin{tabular}[c]{@{}c@{}}46.92\\ \cgray{$^{43.44/44.75/52.59}$}\end{tabular} &
  \begin{tabular}[c]{@{}c@{}}29.90\\ \cgray{$^{29.58/28.26/31.87}$}\end{tabular} &
  \begin{tabular}[c]{@{}c@{}}33.75\\ \cgray{$^{30.18/29.59/41.49}$}\end{tabular} &
  \begin{tabular}[c]{@{}c@{}}41.70\\ \cgray{$^{38.81/43.11/43.19}$}\end{tabular} &
  \begin{tabular}[c]{@{}c@{}}51.82\\ \cgray{$^{51.71/51.88/51.88}$}\end{tabular} &
  \begin{tabular}[c]{@{}c@{}}78.61\\ \cgray{$^{77.94/79.71/78.17}$}\end{tabular} \\

\SYS{} W4.4A8 &
  \begin{tabular}[c]{@{}c@{}}48.17\\ \cgray{$^{46.27/52.58/45.66}$}\end{tabular} &
  \begin{tabular}[c]{@{}c@{}}30.09\\ \cgray{$^{29.17/31.75/29.36}$}\end{tabular} &
  \begin{tabular}[c]{@{}c@{}}34.53\\ \cgray{$^{31.08/43.26/29.26}$}\end{tabular} &
  \begin{tabular}[c]{@{}c@{}}41.74\\ \cgray{$^{39.32/44.79/41.11}$}\end{tabular} &
  \begin{tabular}[c]{@{}c@{}}52.70\\ \cgray{$^{53.67/51.96/52.47}$}\end{tabular} &
  \begin{tabular}[c]{@{}c@{}}79.00\\ \cgray{$^{78.20/78.58/80.21}$}\end{tabular} \\

\SYS{} W8A8 &
  \begin{tabular}[c]{@{}c@{}}48.84\\ \cgray{$^{46.84/54.35/45.34}$}\end{tabular} &
  \begin{tabular}[c]{@{}c@{}}30.93\\ \cgray{$^{30.51/33.21/29.07}$}\end{tabular} &
  \begin{tabular}[c]{@{}c@{}}35.54\\ \cgray{$^{33.00/43.80/29.83}$}\end{tabular} &
  \begin{tabular}[c]{@{}c@{}}40.94\\ \cgray{$^{37.32/44.91/40.59}$}\end{tabular} &
  \begin{tabular}[c]{@{}c@{}}52.10\\ \cgray{$^{53.24/50.94/52.13}$}\end{tabular} &
  \begin{tabular}[c]{@{}c@{}}79.42\\ \cgray{$^{78.98/78.88/80.40}$}\end{tabular} \\ \bottomrule
\end{tabular}
\end{table*}

\subsection{Downstream Tasks Evaluation}
\label{subsec:downstream-eval}

\cb{
We first evaluate GSM8K~\cite{cobbe2021gsm8k} on Qwen2.5 7B model, to validate the quantization efficacy on long-reasoning tasks.
Using \SYS{} W4.4A8 quantization, the \texttt{strict-match} metric only drops from 0.8 (the float16 model) to 0.792, only a 0.008 drop.
Besides this, we evaluate a large number of the downstream tasks on three popular LLMs, shown in Tab.\ref{tab:zero-shot}.
}
The result shows that:
\textbf{1)} \SYS{} W4.4A8 outperforms all the 4-bit weight quantizations, with only 10\% more bit-width.
For example, for the MMLU-Pro task, the average metric of \SYS{} W4.4A8 is improved by 1.69, 6.93, and 0.93 over QServe, QuaRot W4A4, and QuaRot W4A8, respectively.
\textbf{2)} \SYS{} W8A8 is nearly lossless, showing higher accuracy than SmoothQuant.
This comes from the group-wise quantized activation of \SYS{}.


\subsection{Ablation Study}
\label{subsubsec:ablation}

\begin{figure}[]
    \centering
    \includegraphics[width=\columnwidth]{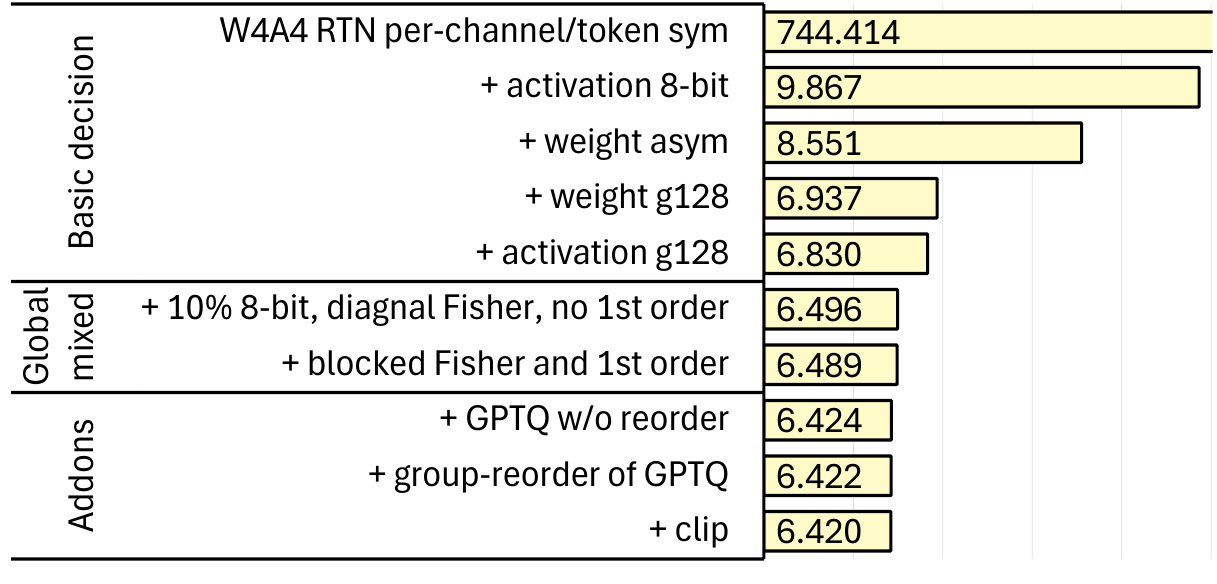}
    \caption{The perplexity (wikitext2) of Llama 3.1 8B model with different configurations.}
    \label{fig:ablation}
\end{figure}

Fig.\ref{fig:ablation} shows the perplexity of Llama 3.1 8B model by adding different optimizations gradually.
With the basic RTN quantization, using 8-bit for activation, and asymmetric and group-wised weight quantization contribute significantly to the accuracy improvement.
This demonstrates the effectiveness of the decisions made in Sec.\ref{subsubsec:configuration-decision}.
Based on these decisions, the 10\% of 8-bit output features improves the accuracy to a high level, for which using blocked Fisher and including the first-order Taylor factor also contributes to the accuracy.
Finally, applying GPTQ and clipping can further improve the accuracy.

\begin{table*}[ht]
\center
\scriptsize
\caption{PPL (wikitext2) comparison with the reported numbers in the related works.}
\label{tab:ppl-more}
\begin{tabular}{@{}cccccccccc@{}}
\toprule
Model &
  FP16 &
  \begin{tabular}[c]{@{}c@{}}GPTQ\\ W4A16\end{tabular} &
  \begin{tabular}[c]{@{}c@{}}AWQ\\ W4A16\end{tabular} &
  \begin{tabular}[c]{@{}c@{}}SqueezeLLM\\ W4A16 0.45\%\end{tabular} &
  \begin{tabular}[c]{@{}c@{}}OmniQuant\\ W4A16/W4A4\end{tabular} &
  \begin{tabular}[c]{@{}c@{}}AfineQuant\\ W4A16/W4A4\end{tabular} &
  \begin{tabular}[c]{@{}c@{}}Atom\\ W4A4 128 outliers\end{tabular} &
  \begin{tabular}[c]{@{}c@{}}SpinQuant\\ W4A8\end{tabular} &
  \begin{tabular}[c]{@{}c@{}}\SYS{}\\ W4.4A8\end{tabular} \\ \midrule
LLaMA 2 7B &
  5.47&
  5.59 &
  5.60 &
  5.57 &
  5.58/14.26 &
  5.58/12.69 &
  6.03 &
  5.7 &
  5.55 \\
LLaMA 3 8B &
  6.14 &
  6.46 &
  6.55 &
  - &
  - &
  - &
  7.57 &
  6.5 &
  6.32 \\ \bottomrule
\end{tabular}
\end{table*}

\subsection{Comparison with More Related Work}
\label{subsec:compare-more-related}

\begin{table}[]
\center
\scriptsize
\caption{PPL of Atom and \SYS{} with similar bit-width.}
\label{tab:compare-atom}
\begin{tabular}{@{}ccc@{}}
\toprule
Models & Llama 2 7B & Llama 2 13B \\
\midrule
Atom W4.4A8 & 5.64 & 5.03 \\
\SYS{} W4.4A8 & 5.54 & 4.93 \\
\bottomrule
\end{tabular}
\end{table}

\textbf{Comparison with Atom.}
Tab.\ref{tab:compare-atom} shows the perplexity of Atom and \SYS{} with the similar bit-width (i.e., W4.4A8).
We use 512 outliers in Atom so its weight is around 4.4-bit on average (using Atom's opensourced code with git commit 7e3618b).
It shows that \SYS{} has much better accuracy than Atom.
We also evaluated the kernel performance of Atom, showing that our W4.4A8 kernel achieves 1.56x speedup than Atom’s W4A8 kernel for the linear layer with sequence length 1024. This means our proposed memory and computation pipeline has much better performance than the related mixed-precision work.

\textbf{Comparison with SliM-LLM.}
Tab.\ref{tab:compare-slim-llm} shows the perplexity of SliM-LLM W4A16 (it only supports weight-only) and \SYS{} with different bit-width.
SliM-LLM W4A16 is evaluated using the opensourced code.
It is the mix of 3/4/5-bits in its code, and the precision is searched within each single layer, rather than using the global precision search.
Note that \SYS{} can support the mix of any bit-width.
We evaluated \SYS{} with the mix of 4-bit/6-bit (W4.2A8, 4.2-bit on average with 90\% 4-bit and 10\% 6-bit), and the mix of 3-bit/4-bit/6-bit (W3.9A8, 3.9-bit on average with 30\% 3-bit, 60\% 4-bit, and 10\% 6-bit) in this table.
We notice even \SYS{} W4A8 (uniform 4-bit for the weight) can defeat SliM-LLM W4A16. This is understandable because SliM-LLM itself focuses on 2-bit and 3-bit optimization, and its paper's main body does not show any 4-bit results.
In contrast, \SYS{} W3.9A8 can defeat \SYS{} W4A8. This demonstrates that \SYS{}'s global precision search can better allocate bit-width to the important weight elements to achieve the nearly lossless results.
Note that the percentage of bit-widths is determined intuitively here, so that it may not be the optimal percentage to achieve the optimal accuracy. How to determine the portion of different bit-width can be a new research problem.

\begin{table}[]
\center
\scriptsize
\caption{PPL (wikitext2) of SliM-LLM and \SYS{}.}
\label{tab:compare-slim-llm}
\begin{tabular}{@{}cccccc@{}}
\toprule
\multirow{2}{*}{Models} & \multicolumn{2}{c}{Llama} & \multicolumn{3}{c}{Qwen2.5} \\
 & 3.2 1B & 3.1 8B & 0.5B & 1.5B & 7B   \\
\midrule
SliM-LLM W4A16 & 10.80 & 6.55 & 14.85 & 9.68 & 7.02 \\
\SYS{} W3.9A8 & 10.20 & 6.50 & 13.68 & 9.51 & 6.97 \\
\SYS{} W4A8 (uniform) & 10.36 & 6.54 & 14.43 & 9.66 & 7.03 \\
\SYS{} W4.2A8 & 10.07 & 6.43 & 13.53 & 9.45 & 6.93 \\
\SYS{} W4.4A8 & 10.05 & 6.42 & 13.42 & 9.44 & 6.92 \\
\bottomrule
\end{tabular}
\end{table}


We compare \SYS{} with more recent quantization works according to the reported numbers in their papers (Tab.\ref{tab:ppl-more}), showing that \SYS{} achieves superior accuracy to a broad range of related works with similar memory consumption.

\subsection{Overhead of Global Precision Search}
\label{subsec:overhead}

\begin{table}[]
\center
\scriptsize
\caption{The overhead of global precision search.}
\label{tab:overhead}
\begin{tabular}{@{}cccccc@{}}
\toprule
\multirow{2}{*}{Models} & \multicolumn{2}{c}{Llama 3.1} & Mistral & \multicolumn{2}{c}{Qwen2.5} \\
                           & 8B           & 70B               & 7B v0.3 & 1.5B    & 7B    \\ \midrule
Time (min) & 7            & 55       & 7       & 2        & 7  \\ 
\bottomrule
\end{tabular}
\end{table}

Tab.\ref{tab:overhead} shows the global precision search overhead described in Sec.\ref{subsec:search}.
As noted in Sec.\ref{subsec:setup}, the calibration dataset has 128 samples with sequence length of 2048.
We use a single A100 GPU for the 1.5B, 7B and 8B models, and 4 A100 GPUs for the 70B models to perform the search.
We make use of \texttt{device\_map} in huggingface for multi-GPU execution, which is sequential execution of layers on different devices.
The 7B/8B models require about 7 minutes and the 70B models require less than 60 minutes to complete the search.
Considering that the quantization only needs to be performed once, the searching algorithm is practical for the real workloads.
In contrast, SliM-LLM takes more than 3 hours to determine the precision of Llama 3.1 8B model and more than 4 hours for Qwen2.5 7B model according to our experiments.

\subsection{High Precision Distribution}
\label{subsec:precision-distribution}

Fig.\ref{fig:percentage-of-8bit} shows the percentage of 8-bit out features in each of the linear layers of Llama 3.1 8B, with 10\% global 8-bit out features searched by \SYS{} (i.e., W4.4A8).
It shows that high-salient (i.e., 8-bit) features are distributed very differently in different linear layers.
Specifically, the \texttt{v\_proj} and \texttt{down\_proj} layers show much higher percentage of high-salient features than other layers,
for which Tab.\ref{tab:avg-percentage} shows the average percentage of different classes of linear layers.

\begin{table}[]
\center
\scriptsize
\caption{The average percentage of 8-bit out features in the seven classes of linear layers in Llama 3.1 8B, with 10\% global 8-bit out features in \SYS{}.}
\label{tab:avg-percentage}
\begin{tabular}{@{}cccccccc@{}}
\toprule
layer (xx\_proj)      & q & k & v & o & gate & up & down \\ \midrule
avg 8-bit (\%) & 3.93    & 12.36   & 71.22   & 18.70   & 0.73       & 1.46     & 53.82      \\ \bottomrule
\end{tabular}
\end{table}

\subsection{One-pass vs. Progressive Search}
\label{subsec:one-pass-eval}

As described in Sec.\ref{subsec:search}, \SYS{} searches the high-salience features within a single pass rather than iteratively identifying the high-salience parts in a smaller step, as we observe the single-pass method show similar results with the iterative method and saves a lot of computation overhead than the latter.
We have tried the progressive procedure on Llama 3.1 8B and Mistral 7B models, which identifies smaller portions of the high-salience features iteratively.
Results show that the accuracy is the same to the one-pass method to two decimal places.
However, the progressive method shows much higher search time due to the repeated procedure.
The one-pass method takes 7 minutes for each of the two models to search 10\% high-salience features,
while the progressive method that searches 2\% high-salience iteratively takes 30 minutes to find top 10\% features.


\subsection{Working with KV Quantization}
\label{subsec:with-kv-quant}

\begin{table}[]
\center
\scriptsize
\caption{PPL of enabling KV Quantization in \SYS{}.}
\label{tab:with-kv-quant}
\begin{tabular}{@{}ccc@{}}
\toprule
Models & Llama 3.1 8B & Qwen2.5 7B \\
\midrule
\SYS{} W4.4A8 w/o KV quant & 6.42 & 6.92 \\
\SYS{} W4.4A8 KV8          & 6.42 & 6.96 \\ 
\bottomrule
\end{tabular}
\end{table}

It is straightforward to apply any KV quantization technology together with \SYS{}’s weight-activation quantization, as KV quantization is orthogonal to the weight-activation quantization.
Tab.\ref{tab:with-kv-quant} shows the perplexity (wikitext2) of Llama 3.1 8B and Qwen 2.5 7B for enabling and disabling KV quantization.
It shows that the 8-bit KV quantization is nearly lossless upon \SYS{}.

\subsection{Calibration Dataset for Salience Search}
\label{subsec:calib-size}

We use 128 samples for the calibration to search the high-salience channels in \SYS{}, which is a common configuration in the existing solutions.
We also find that a smaller dataset can still identify the high-salience channels in \SYS{}.
The perplexity of using different samples for \SYS{} salience search are shown in Tab.\ref{tab:num-samples}.
This shows that shrinking the sample size from 128 to 16 for the global salience search has negligible effect to the accuracy.

\begin{table}[]
\center
\scriptsize
\caption{PPL (wikitext2) when using different number of samples for salience search (W4.4A8 quantized).}
\label{tab:num-samples}
\begin{tabular}{@{}ccccc@{}}
\toprule
\#samples & 128 & 64 & 32 & 16 \\
\midrule
Llama 3.1 8B & 6.42 & 6.42 & 6.42 & 6.42 \\
Qwen2.5 7B & 6.92 & 6.92 & 6.92 & 6.93 \\
\bottomrule
\end{tabular}
\end{table}

We also use different dataset families between salience searching and perplexity evaluation to evaluate the effect of input distribution shift.
When using c4 dataset for salience search and evaluate the perplexity on wikitext2 (W4.4A8 quantized), 
the Llama 3.1 8B and Qwen2.5 7B has the same perplexity with that using wikitext2 for the calibration.
This shows that changing the calibration dataset does not affect the accuracy for the same task.
In another word, when the data distribution between the calibration dataset and real input is different, \SYS{}'s salience search algorithm still works well.

\section{Summary}
\label{sec:summary}

We have presented \SYS{}, achieving high accuracy with low memory consumption and high system efficiency with the rarely explored optimization space of mixed-precision quantization between output features.
\SYS{} identifies the salience of each output feature according to the loss w.r.t. the global model rather than each single layer.
By assigning larger bit-width to the features need it the most, \SYS{} achieves the superior accuracy to SOTA with low memory consumption.
The sub-problems of different bit-widths are disjoint and can run in parallel efficiently on the GPU.
We have identified the sweet spot of the quantization configuration that is friendly to both accuracy and system efficiency.
To address the challenge of system efficiency, we design the two-step dequantization to enable using int8 Tensor Core and the fast integer-float conversion to reduce the dequantization overhead.
We have designed the end-to-end software pipeline to overlap the memory access, the dequantization computation with SIMT Core and the MatMul with Tensor Core to the best.
Experiment results show that \SYS{} achieves superior accuracy to existing works and state-of-the-art system efficiency with low memory cost.

\bibliography{refs}
\bibliographystyle{mlsys2026}


\end{document}